# YOLORe-IDNet: An Efficient Multi-Camera System for Person-Tracking

Vipin Gautam, Shitala Prasad, and Sharad Sinha

School of Mathematics and Computer Science
Indian Institute of Technology Goa
vipin2113106@iitgoa.ac.in, shitala@iitgoa.ac.in, sharad@iitgoa.ac.in

**Abstract.** The growing need for video surveillance in public spaces has created a demand for systems that can track individuals across multiple cameras feeds in real-time. While existing tracking systems have achieved impressive performance using deep learning models, they often rely on pre-existing images of suspects or historical data. However, this is not always feasible in cases where suspicious individuals are identified in real-time and without prior knowledge. We propose a person-tracking system that combines correlation filters and Intersection Over Union (IOU) constraints for robust tracking, along with a deep learning model for cross-camera person re-identification (Re-ID) on top of YOLOv5. The proposed system quickly identifies and tracks suspect in real-time across multiple cameras and recovers well after full or partial occlusion, making it suitable for security and surveillance applications. It is computationally efficient and achieves a high F1-Score of 79% and an IOU of 59% comparable to existing state-of-the-art algorithms, as demonstrated in our evaluation on a publicly available OTB-100 dataset. The proposed system offers a robust and efficient solution for the real-time tracking of individuals across multiple camera feeds. Its ability to track targets without prior knowledge or historical data is a significant improvement over existing systems, making it well-suited for public safety and surveillance applications.

## 1 Introduction

Visual security systems have become increasingly important in today's world from a surveillance and law and order point of view. Advancements in technology, such as high-definition images and artificial intelligence (AI) can be leveraged to create robust person and object identification and tracking systems. Traditional security approaches involve monitoring by human personnel, which can be challenging for large areas like malls, smart cities, and universities. Surveillance cameras are commonly used in public places, but it is difficult to compare different individuals manually. AI-powered surveillance systems [1] have been developed to monitor suspects in real-time across multiple cameras, addressing these challenges.

Technically, intelligent surveillance systems [18] can be divided into two parts: intra-camera tracking and inter-camera tracking, which employs recognition strategies. There are various challenges in both strategies, such as viewpoint variation, occlusion, different aspect ratios and spatial sizes, lighting, and cluttered background, to name a few. In the case of intra-camera tracking, the difficulty is due to occlusion, which normally happens when people overlap each other or are overlapped by some other object, which leads a tracking algorithm to track the wrong target. In cases of inter-camera tracking, the difficulty is due to high intra-class variance due to variations in lighting conditions, angles, resolutions, and other factors. In addition to this, the task becomes more challenging when we have a large number of people to identify and a large number of non-overlapping cameras, as shown in Fig. 1 with high intra-class variance.

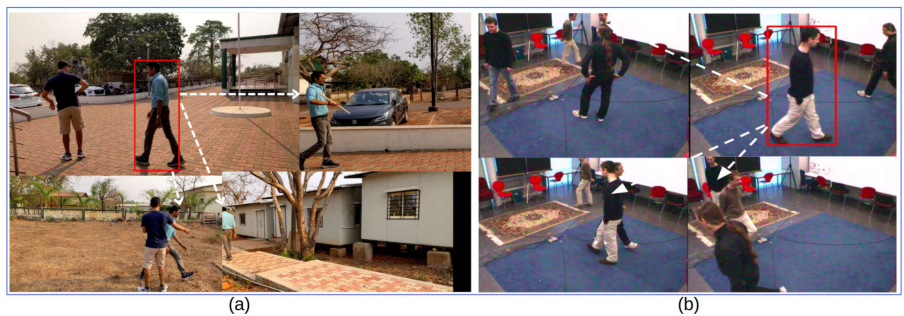

Fig. 1: Person tracking in a real-world scenario, showcasing the challenges of high intra-class variability: (a) MCPT Dataset and (b) EPFL dataset.

In this work, we present YOLORe-IDNet. The YOLORe-IDNet combines Kernelized Correlation Filters (KCF) [7] with an IOU constraint and You Only Look Once (YOLOv5) [9] to form a more robust and efficient tracker for intra-camera tracking tasks. In order to reduce tracking errors and improve accuracy, an IOU-based occlusion assessment approach is introduced into our framework. For recognizing a person across multiple camera views, a deep learning based person Re-ID model has been used along with camera network information to reduce the latency. We make the following novel contributions:

- Creation of an open dataset using cameras placed at 9 different locations in the university campus (see Fig. 6), where the dataset consists of 81000 frames.
- User defined real-time identification and selection of person of interest in a camera feed by a simple mouse gesture.
- A novel algorithm for intra-camera tracking that uses IOU-based novel occlusion assessment approach to effectively address occlusion and minimize tracking failures.

The rest of the paper is organized as follows: the related work on visual object tracking for intra-camera tracking and inter-camera tracking is presented in section 2, and section 3 elaborates details of the methodology presented in this paper. Section 4 discusses the experimental setup and results using our dataset and comparison with some of the state-of-the art tracking algorithms. Finally, in section 5, we conclude the paper and discuss potential directions for future work.

## 2 Related Work

The field of computer vision (CV) has played a significant role in developing intelligent surveillance systems. Person tracking, in particular, is a critical component of these systems. In this section, we review the existing work on person-tracking methods and their strengths and weaknesses. Compared to conventional computer vision approaches, modern deep learning models have achieved significant progress in person tracking due to their ability to learn features and patterns in a data-driven manner. However, the existing models often require a large amount of labeled data for training and can be computationally expensive.

### 2.1 Intra-camera tracking

Intra-camera object tracking refers to the process of tracking objects within a single video feed. Correlation filters [5] have been used for such a tracking exercise. These filters work by correlating a template with the target object in each frame of a video. One of the earliest approaches for single camera tracking used [4] MOSSE tracker. However, these trackers can be susceptible to distractors and are limited in their ability to handle occlusion and scale adaptation. Intersection over union (IOU) constraint-based tracking [3] is another approach that has been used to track individuals in real-time scenarios. This approach computes the IOU between the current frame and previous frames to predict the target's location. However, IOU-based tracking can be limited in its ability to handle complex scenarios with multiple objects.

The combination of a CNN based object detector and correlation filter can significantly enhance the robustness and tracking accuracy [17]. The detectors can quickly identify potential targets in a given scene, despite occlusions and scale variations. On the other hand, correlation filters are effective for tracking objects with stable appearances and minimal scale change but struggle when objects undergo significant changes in appearance or shape. By integrating the strengths of both detectors and correlation filters, it is possible to leverage their complementary strengths and achieve improved tracking performance in challenging scenarios [17]. Our proposed approach incorporates YOLOv5 and correlation filter techniques, along with an occlusion detection module, to effectively mitigate tracking drifts. The selection of YOLOv5 and correlation filter, and the design of the occlusion detection module are done keeping in view real-time processing requirements.

### 2.2   Inter-camera tracking

Inter-camera tracking tracks objects across multiple cameras, which is a challenging problem due to the differences in camera viewpoints, image quality, illumination conditions etc. In recent years, significant progress has been made in developing inter-camera tracking methods that are robust to these challenges. Several studies have investigated the use of camera networks for human tracking, including the review [8] on various tracking techniques and challenges. One common approach to inter-camera tracking is to use appearance-based methods such as person Re-ID [15]. Re-ID is a process of matching people across different cameras by using their appearance features such as color, texture, and shape. These models typically learn discriminative feature representations of people that are invariant to pose, illumination, and camera viewpoint changes. Therefore, we base our inter-camera tracking module upon AlignedReID++ [14] a person Re-ID model. It uses DMLI, a dynamically matching local information method, for person Re-ID and is effective at resolving pose misalignments or other challenging samples.

Another important aspect of inter-camera tracking is the use of camera network information. This includes the spatial relationships, the field of view (FOV), and the calibration parameters of each camera. By incorporating this information, inter-camera tracking methods can better estimate the trajectories of people across different cameras.

## 3   YOLORe-IDNet: Proposed Methodology

In this section we first present the algorithms in our methodology and then the implementation of the algorithms for real-time processing.

### 3.1   Proposed Algorithm

YOLORe-IDNet reads the Real Time Streaming Protocol (RTSP) URLs or paths of the video sources from a "txt" file. Upon identifying the target, the user of the system would input the camera number in the ROI selection field and be expected to draw a bounding box around the person who needs to be tracked. The system then encodes the target data and sends it to the cloud server via post request along with the camera number where the target was first seen. The cloud server then processes this request and validates whether or not the inter-camera tracking module needs to be activated, as illustrated in Fig. 2.

Inside the intra-camera tracking module, we first validate if this is the first request made to the server. If so, then the system performs feature extraction with the help of YOLO detections, finds a similarity score, and initializes the tracker with the coordinates of the person whose similarity score is highest with the target. This ensures the current location of the target is obtained since in real-time systems, Due to potential latency in the network and processing delays, the target may have moved from its original location by the time it is found.

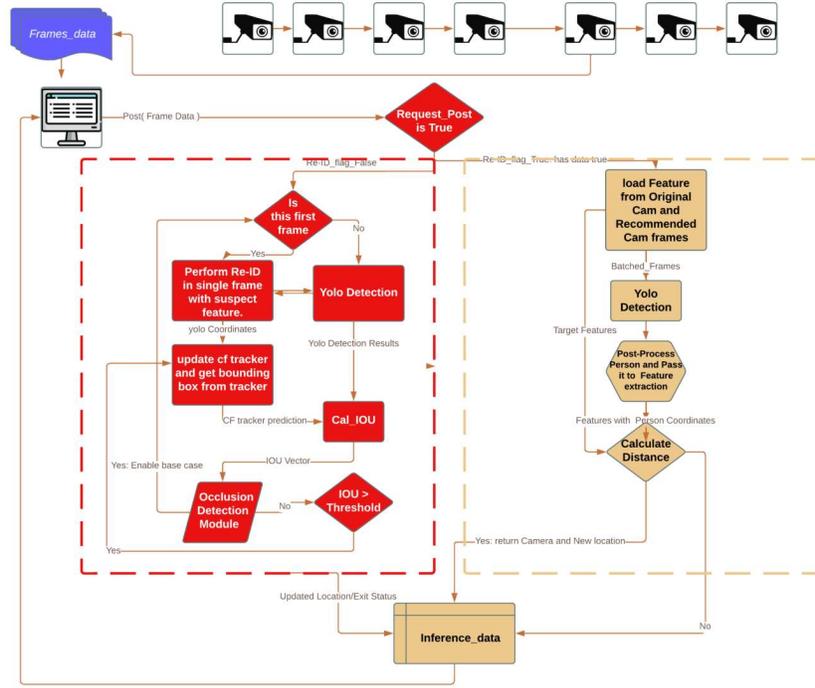

Fig. 2: Flow diagram with intra-camera tracking on left and inter-camera tracking on right.

Similarly, when the person exists from the camera's FOV, the cloud server sends the trigger to the client, and from there onwards the inter-camera tracking begins. In this way, the system continues to track the target and finally saves the trajectory map of the locations where the target visited, as shown in Fig. 6.

**Intra-camera tracking** The intra-camera tracking module consists of two stages. First, a base case is activated when the module is called for the first time or when occlusion is observed by the system in the current stream, as mentioned in the proposed Algorithm 1 (line 2). In the second stage of algorithm (line 9), bounding box detections are obtained using YOLOv5 [9]. The correlation filter updates the target's new location, over which an IOU constraint is applied to select the bounding box with the highest IOU compared to a threshold of 30%, assuming that the movement of the target is limited. The system then applies the proposed Occlusion Detection Algorithm (Algorithm 2) to further assess whether the target has been occluded to overcome tracking errors. This reduces overall tracking errors that occur when the target is fully or partially occluded by other objects or people. If the occlusion is detected, the base case is activated. The features of the target are extracted along with the features of all the detected persons in the frame, and Re-ID is applied. The tracker is then updated with the coordinates of the person where the highest similarity is observed. Otherwise, the system continues to track the target. Fi-nally, the tracking module returns the target's location to the client, and the target's current location, along with the camera ID, is stored to update the target's trajectory. The resulting trajectory map is stored as an output in Fig 6.

---
**Algorithm 1:** Intra-camera tracking algorithm
---
**Input:** Image_frame, target
**Output:** yolo_bbox

1 $base\_case \leftarrow True$;
2 **if** $base\_case$ is True **then**
3     bboxes ← run_yolo_detector(Image_frame); base_case = False;
4     objects_features = extract_features(bboxes, Image_frame);
5     target_features = extract_features(target);
6     target_new_coords = perform_reid(objects_features, target_features);
7     initialize_tracker(target_new_coords, Image_frame);
8     **return** target_new_coords
9 **else**
10     bboxes = run_yolo_detector(Image_frame);
11     target_new_coords = update_tracker(Image_frame);
12     IOU_vector, yolo_bbox, has_exit = apply_iou_constraint(bboxes, target_new_coords);
13     **if** not has_exit **then**
14        **if** $max\_of(IOU\_vector) \geq Threshold$ **then**
15           occlusion_flag = detect_occlusion(IOU_vector);
16           **if** occlusion_flag is True **then**
17              base_case = True;
18           **else**
19              initialize_tracker(yolo_bbox, Image_frame);
20              **return** yolo_bbox;
21     **else**
22        has_exit;

---
**Algorithm 2:** Occlusion Detection Algorithm
---
**Input** : IOU_vector
**Output:** Boolean value indicating occlusion

1 $count \leftarrow 0$; $max\_iou \leftarrow$ max(IOU_vector);
2 **for** $iou\_score$ in $IOU\_vector$ **do**
3     **if** $iou\_score > 0.1$ and $iou\_score \neq max\_iou$ **then**
4        $count \leftarrow count + 1$;
5 **if** $count > 1$ **then**
6     **return** True;
7 **else**
8     **return** False;

To establish the exit of the target from the current stream, the IOU vector of the last three frames is observed. If the IOU vector is zero in the last three frames, the target is marked as exited, triggering the inter-camera tracking module. The system output shown in Fig. 3a demonstrates the performance of our occlusion detection algorithm on both our MCPT dataset and the EPFL dataset's test sequence [6]. Specifically, in subfigures (A) and (B), the targets were occluded in the second frame, but our algorithm was able to detect this and successfully recovered tracking from the third frame onwards.

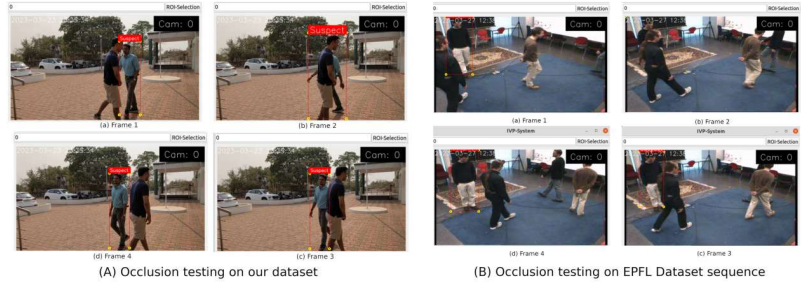

(a) Demonstration of occlusion handling

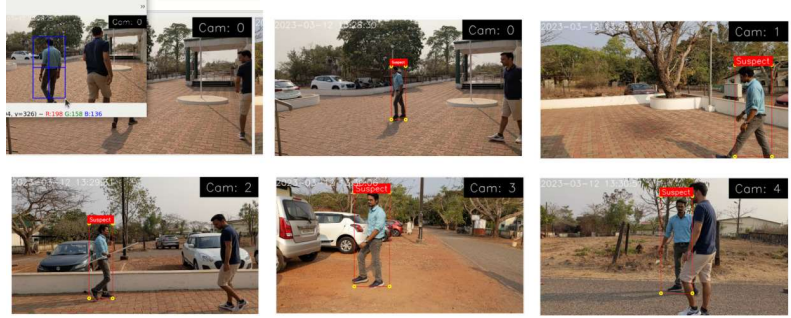

(b) Target location at different intervals: Capturing target movement

Fig. 3: System output on various video sequences.

**Inter-camera tracking** In the proposed work we make use of AlignedReID++ [14] for person Re-ID. The network uses the DMLI method that aligns horizontal stripes in person and learns global and local features. The local branch aligns local parts using the shortest path distance, which helps the global branch learn more discriminative features for robust Re-ID.

The main goal of the inter-camera tracking module is to conduct the search for the target, and after a successful search, return the camera-ID and bounding box coordinates of the target. As shown in Fig.4 in the inter-camera tracking module, initially target features are extracted using the Re-ID model, and the YOLO detector is applied to a batch of frames; subsequently, the bounding boxes are filtered to obtain person detections, which serve as candidates for Re-ID. Furthermore, post-processing is done to crop the images of all the persons in the frames, and another batch is formed and sent for Re-ID where the similarity match is performed with the original target features. In cases of success, the camera-ID and location are returned to the client; otherwise, the system continues the search with recommended cameras in the next iteration.

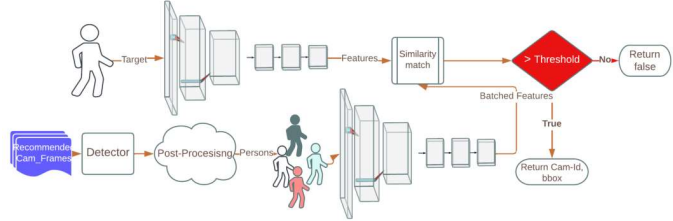

Fig. 4: Inter camera tracking flow diagram.

### 3.2 System Architecture and Implementation

**Multithreading for Enhancing Execution Speed** To improve the client's performance during the I/O-blocking operation of reading from streaming cameras, we employed a multi-threading approach. Instead of relying on a single thread to grab frames sequentially and risking delays, we spawned an additional thread to handle grabbing frames in parallel. This enabled the continuous reading of frames from the I/O thread while allowing the root thread to process the current frame. In essence, by creating another thread to grab frames, we could enhance the execution of the client.

**REST APIs for inference Data** A REST API (Representational State Transfer API) is an architectural style for building web services. REST APIs communicate over HTTP requests.

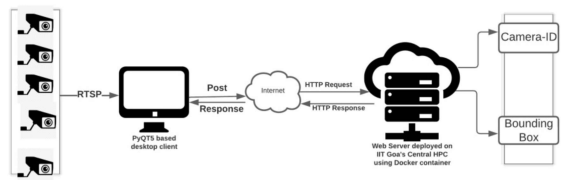

Fig. 5: System architecture for cloud-based application

The complete architecture of the deployed system is shown in Fig. 5. The system architecture is based on a Flask server that receives video frames through HTTP requests. These frames are processed using object detection and tracking algorithms, as explained earlier. Once the cloud server generates inference data, a JSON response is prepared by the server, including bounding boxes and stream ID for the suspect. This response is then relayed back to the client as an HTTP response. Finally, the inference data is visualized on the client, allowing the generation of a target trajectory, as demonstrated in Fig. 3b and 6.

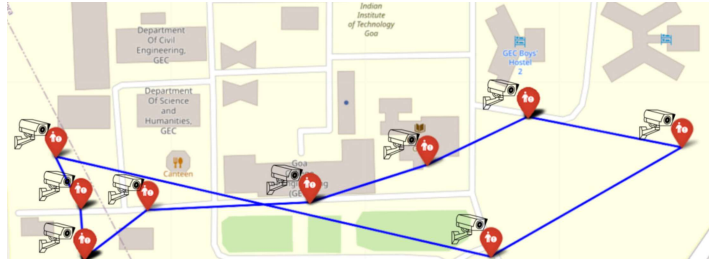

Fig. 6: Trajectory map saved by system as output

**Multi-Camera Person Tracking (MCPT) Dataset** In general, person-tracking systems need datasets of the target's trajectory to be developed and evaluated. However, there is a shortage of such datasets that cover multiple cameras, making it difficult to evaluate tracking algorithms against obstacles like occlusions, scale and angle variations, and the movement of individuals in different zones.

To address this challenge, we created the MCPT dataset to evaluate person-tracking systems in real-world scenarios. The dataset consists of nine videos captured from non-overlapping cameras situated across the university campus, each of which is approximately 5 minutes long, resulting in a total of approximately 81,000 frames. Each camera was set to a resolution of 720 x 1280 and recorded at a frame rate of 30 frames per second (FPS). To generate suspicious trajectories that are representative of real-world scenarios, a target was recorded in each stream for a maximum of 30 seconds. The dataset captured the target amidst various intra-class variations, such as targets with varying angles, scale, and lighting variations, while also including instances of full and partial occlusion, providing a more realistic representation of real-world scenarios. The dataset is provided in a standard format, with each video stored as a sequence of frames in JPEG format and each frame labeled with the corresponding ground truth data in a separate "txt" file.

## 4 Experimental Results

**Datasets and Evaluation Protocol:** We evaluated the YOLORe-IDNet on MCPT dataset and Object Tracking Benchmark-100 (OTB-100) dataset. The OTB-100 [16] dataset is a widely utilized benchmark dataset that is primarily used to evaluate the performance of visual object tracking algorithms. This dataset is composed of 100 video sequences, which present a wide range of visual challenges, including occlusion, motion blur, changes in lighting, scale variation, and background clutter. Moreover, this dataset features a diverse set of object classes, including humans, animals, vehicles, and other miscellaneous objects.

Given that the primary objective of the YOLORe-IDNet is to track individuals, we deliberately selected 11 of the most challenging person sequences from this dataset for our evaluation.

Precision, Recall, F1Score, Intersection over union (IOU), and Overall precision error (OPE) are used as evaluation metrics.

$$Precision = \frac{TP}{TP+FP}, \ Recall = \frac{TP}{TP+FN}, \ F1Score = 2*\frac{Precision*Recall}{Precision+Recall} \quad (1)$$

$$OPE = \frac{1}{N}\sum_{i=1}^{N} d_i, \ IOU = \frac{|GT \cap PD|}{|GT \cup PD|} \quad (2)$$

where $N$ is the total number of matches for ground truth, $d_i$ is the Euclidean distance between the center of the predicted bounding box ($PD$) and the center of the ground truth bounding box ($GT$) in the $i^{th}$ frame. $IOU$ is the overlap between the $PD$ and the $GT$ box at frame f. True positives ($TP$) refer to the number of correctly identified $GT$ matches, while false positives ($FP$) refer to the number of predicted matches that do not match $GT$. False negatives ($FN$) refer to the number of $GT$ matches that are not identified by the tracker.

**Evaluation on MCPT dataset:** As shown in Fig. 3b, the system is able to accurately track the targets in real-time. To assess the performance of our method, we computed Precision, Recall, and mean IOU for each camera, as reported in Table 1. During the experiments, the proposed system maintained an operational speed of 18 FPS, showcasing its capability to handle real-time feeds efficiently. Moreover, our method demonstrated high Precision and IOU values of 100% and 91% respectively, indicating its ability to provide accurate object localization across different camera views.

Table 1: Results on MCPT dataset

|  | Cam0 | Cam1 | Cam2 | Cam3 | Cam4 | Cam5 | Cam6 | Cam7 | Cam8 | Mean |
|---|---|---|---|---|---|---|---|---|---|---|
| **Precision↑** | 1.00 | 1.00 | 1.00 | 1.00 | 1.00 | 1.00 | 1.00 | 1.00 | 1.00 | 1.00 |
| **Recall↑** | 0.82 | 0.81 | 0.78 | 0.70 | 0.75 | 0.91 | 0.73 | 0.93 | 0.85 | 0.81 |
| **IOU↑** | 0.91 | 0.90 | 0.89 | 0.93 | 0.92 | 0.92 | 0.91 | 0.92 | 0.90 | 0.91 |

Note: FPS for all camera feeds = 18.

**Ablation study:** As shown in Table 2 the ablation study was conducted to evaluate the effectiveness of the occlusion assessment module in improving the performance of the system. The results indicate that the inclusion of the occlusion assessment module led to a notable increase of 6% in F1Score and 5% in Recall, and 33.73 units decrease in OPE when compared to the system without the occlusion assessment module. The reason for this is that, when a target occludes, the tracker continues to track the wrong target in the absence of occlusion assessment and without any self-correcting mechanism.

Table 2: Ablation study on Occlusion assessment module.

| | With occlusion assessment | | | Without occlusion assessment | | |
|---|---|---|---|---|---|---|
| | F1Score↑ | Recall↑ | OPE↓ | F1Score↑ | Recall↑ | OPE↓ |
| Cam0 | 0.90 | 0.82 | 6.14 | 0.77 | 0.78 | 103.39 |
| Cam1 | 0.89 | 0.81 | 7.75 | 0.90 | 0.82 | 7.22 |
| Cam2 | 0.87 | 0.78 | 6.58 | 0.88 | 0.78 | 5.87 |
| Cam3 | 0.82 | 0.70 | 5.68 | 0.85 | 0.74 | 5.98 |
| Cam4 | 0.86 | 0.75 | 6.25 | 0.70 | 0.63 | 115.4 |
| Cam5 | 0.95 | 0.91 | 6.21 | 0.77 | 0.72 | 104.8 |
| Cam6 | 0.85 | 0.73 | 6.12 | 0.87 | 0.78 | 5.84 |
| Cam7 | 0.96 | 0.93 | 5.94 | 0.87 | 0.78 | 5.55 |
| Cam8 | 0.92 | 0.85 | 5.55 | 0.88 | 0.78 | 5.75 |
| **Mean** | 0.89 | 0.81 | 6.25 | 0.83 | 0.76 | 39.98 |

Upward arrow: Higher is desired; Downward arrow: Lower is desired.

**Comparison with State-of-the-art Methods:** We selected several popular tracking algorithms for comparison, and our method achieved a high F1Score of 79%, second only to CSRT's 83%. However, our method outperformed CSRT's in terms of OPE (10.83 vs. 15.51) on OTB-100 Dataset, as shown in Table 3. Moreover, our algorithm achieved a relatively high mean IOU of 59%. We also compared these tracking algorithms on MCPT dataset, and our approach outperformed other tracking algorithms by a big margin. The reason for this high performance is the ability to recover from occlusions effectively.

Table 3: Comparison with state-of-the-art methods on OTB-100 & MCPT dataset

| Method | OTB-100 Dataset | | | | MCPT dataset | | | |
|---|---|---|---|---|---|---|---|---|
| | F1Score↑ | Recall↑ | OPE↓ | IOU↑ | F1Score↑ | Recall↑ | OPE↓ | IOU↑ |
| Boosting[12] | 0.50 | 0.40 | 53.05 | 0.29 | 0.50 | 0.37 | 198.76 | 0.30 |
| CSRT[13] | **0.83** | **0.82** | 15.51 | **0.61** | 0.33 | 0.27 | 363.89 | 0.23 |
| KCF[7] | 0.39 | 0.29 | 12.38 | 0.22 | 0.23 | 0.16 | 541.20 | 0.12 |
| MFLOW[10] | 0.35 | 0.26 | 65.13 | 0.21 | 0.13 | 0.07 | 319.87 | 0.11 |
| MIL[2] | 0.49 | 0.37 | 51.09 | 0.28 | 0.55 | 0.43 | 146.48 | 0.33 |
| MOSSE[4] | 0.23 | 0.19 | 22.81 | 0.24 | 0.40 | 0.28 | 327.36 | 0.20 |
| TLD[11] | 0.64 | 0.54 | 37.19 | 0.38 | 0.08 | 0.04 | 429.15 | 0.04 |
| **OURS** | 0.79 | 0.69 | **10.83** | 0.59 | **0.89** | **0.81** | **6.25** | **0.91** |

**Bold**: best; Underlined: second-best results.

## 5 Conclusion

In this study, we developed a person-tracking system that uses correlation filters and IOU constraints for reliable tracking and a deep learning model for cross-camera person Re-ID. Our experiments show that the proposed system is effective in detecting and tracking objects in multi-camera surveillance scenarios, achieving an F1Score of 89%, Recall of 81%, and a mean IOU of 91%, with a low OPE of 6.25. We also evaluated our system on the OTB-100 dataset and achieved competitive results with the least OPE of 10.83 and the second-highest F1Score of 79%, Recall of 69%, and mean IOU of 59% compared to state-of-the-art tracking algorithms. In the future, we aim to explore additional techniques, such as apparel invariant Re-ID, to further enhance the accuracy and generalization of our system in real-world scenarios.

## References

1. Ahmed, A.A., Echi, M.: Hawk-eye: An ai-powered threat detector for intelligent surveillance cameras. IEEE Access **9**, 63283–63293 (2021)
2. Babenko, B., Yang, M.H., Belongie, S.: Visual tracking with online multiple instance learning. In: CVPR. pp. 983–990. IEEE (2009)
3. Bochinski, E., Eiselein, V., Sikora, T.: High-speed tracking-by-detection without using image information. 2017 14th IEEE International Conference on Advanced Video and Signal Based Surveillance (AVSS) pp. 1–6 (2017)
4. Bolme, D.S., Beveridge, J.R., Draper, B.A., Lui, Y.M.: Visual object tracking using adaptive correlation filters. In: CVPR. pp. 2544–2550. IEEE (2010)
5. Chen, Z., Hong, Z., Tao, D.: An experimental survey on correlation filter-based tracking. CoRR **abs/1509.05520** (2015), http://arxiv.org/abs/1509.05520
6. Fleuret, F., Berclaz, J., Lengagne, R., Fua, P.: Multicamera people tracking with a probabilistic occupancy map. IEEE trans on PAMI **30**(2), 267–282 (2007)
7. Henriques, J.F., Caseiro, R., Martins, P., Batista, J.: High-speed tracking with kernelized correlation filters. IEEE Trans. on PAMI **37**(3), 583–596 (mar 2015), https://doi.org/10.1109%2Ftpami.2014.2345390
8. Hou, L., Wan, W., Hwang, J.N., Muhammad, R., Yang, M., Han, K.: Human tracking over camera networks: a review. EURASIP **2017**(1), 1–20 (2017)
9. Jocher, G., Chaurasia, A., Stoken, A., Borovec, J., NanoCode012, Yonghye Kwon, E.A.: ultralytics/yolov5: v7.0 - YOLOv5 SOTA Realtime Instance Segmentation (Nov 2022). https://doi.org/10.5281/zenodo.7347926
10. Kalal, Z., Mikolajczyk, K., Matas, J.: Forward-backward error: Automatic detection of tracking failures. In: ICPR. pp. 2756–2759. IEEE (2010)
11. Kalal, Z., Mikolajczyk, K., Matas, J.: Tracking-learning-detection. IEEE trans on PAMI **34**(7), 1409–1422 (2011)
12. Liu, R., Cheng, J., Lu, H.: A robust boosting tracker with minimum error bound in a co-training framework. In: ICCV. pp. 1459–1466. IEEE (2009)
13. Lukezic, A., Vojir, T., Cehovin Zajc, L., Matas, Jiri, a.K.M.: Discriminative correlation filter with channel and spatial reliability. In: CVPR. pp. 6309–6318 (2017)
14. Luo, H., Jiang, W., Zhang, X., Fan, X., Qian, J., Zhang, C.: Alignedreid++: Dynamically matching local information for person re-identification. Pattern Recogn. **94**(C), 53–61 (oct 2019), https://doi.org/10.1016/j.patcog.2019.05.028
15. Ming, Z., Zhu, M., Wang, X., Zhu, J., Cheng, J., Gao, C., Yang, Y., Wei, X.: Deep learning-based person re-identification methods: A survey and outlook of recent works. Image and Vision Computing **119**, 104394 (2022)
16. Wu, Y., Lim, J., Yang, M.H.: Object tracking benchmark. IEEE Trans. on PAMI **37**(9), 1834–1848 (2015). https://doi.org/10.1109/TPAMI.2014.2388226
17. Yang, B., Tang, M., Chen, S., Wang, G., Tan, Y., Li, B.: A vehicle tracking algorithm combining detector and tracker. EURASIP **2020**(1), 1–20 (2020)
18. Zabłocki, M., Gościewska, K., Frejlichowski, D., Hofman, R.: Intelligent video surveillance systems for public spaces–a survey. Journal of Theoretical and Applied Computer Science **8**(4), 13–27 (2014)